\documentclass{article}
\usepackage{acra}
\usepackage{graphicx}
\usepackage{tabularx}
\usepackage{booktabs}
\usepackage{amsmath,amssymb}
\usepackage{todonotes}
\usepackage{subcaption}
\usepackage[nolist,nohyperlinks]{acronym}
\usepackage{tikz,tkz-euclide}
\usepackage[hidelinks]{hyperref}
\usepackage{balance}
\usetikzlibrary{decorations,calligraphy}
\usetikzlibrary{calc}

\UseRawInputEncoding

\begin{acronym}[AAAAAAAAA]
    \acro{1d}[1D]{One-Dimensional}
    \acro{2d}[2D]{Two-Dimensional}
    \acro{3d}[3D]{Three-Dimensional}
    \acro{cas}[CAS]{Centre for Autonomous Systems}
    \acro{cnn}[CNN]{Convolutional Neural Network}
    \acro{cpu}[CPU]{Central Processing Unit}
    \acro{doals}[DOALS]{Urban Dynamic Objects LiDAR}
    \acro{dof}[DoF]{Degree-of-Freedom}
    \acro{dvs}[DVS]{Dynamic Vision Sensor}
    \acrodefplural{dvs}[DVS's]{Dynamic Vision Sensors}
    \acro{ekf}[EKF]{Extended Kalman filter}
    \acro{fifo}[FIFO]{First In, First Out}
    \acro{fov}[FoV]{Field-of-View}
    \acro{gnss}[GNSS]{Global Navigation Satellite System}
    \acrodefplural{gnss}[GNSS's]{Global Navigation Satellite Systems}
    \acro{gp}[GP]{Gaussian Process}
    \acrodefplural{gp}[GPs]{Gaussian Processes}
    \acro{gpm}[GPM]{Gaussian Preintegrated Measurement}
    \acro{ugpm}[UGPM]{Unified Gaussian Preintegrated Measurement}
    \acro{gps}[GPS]{Global Position System}
    \acrodefplural{gps}[GPS's]{Global Position Systems}
    \acro{gpu}[GPU]{Graphic Processing Unit}
    \acro{hdr}[HDR]{High Dynamic Range}
    \acro{hri}[HRI]{Human-Robot Interactions}
    \acro{icp}[ICP]{Iterative Closest Point}
    \acro{idol}[IDOL]{IMU-DVS Odometry using Lines}
    \acro{iou}[IoU]{Intersection over Union}
    \acro{imu}[IMU]{Inertial Measurement Unit}
    \acro{in2laama}[IN2LAAMA]{INertial Lidar Localisation Autocalibration And MApping}
    \acro{kf}[KF]{Kalman Filter}
    \acro{kl}[KL]{Kullback–Leibler}
    \acro{lidar}[LiDAR]{Light Detection And Ranging Sensor}
    \acro{lpm}[LPM]{Linear Preintegrated Measurement}
    \acro{map}[MAP]{Maximum A Posteriori}
    \acro{mle}[MLE]{Maximum Likelihood Estimation}
    \acro{ndt}[NDT]{Normal Distribution Transform}
    \acro{pca}[PCA]{Principal Component Analysis}
    \acro{pm}[PM]{Preintegrated Measurement}
    \acro{rcnn}[R-CNN]{Region-based Convolutional Neural Network}
    \acro{rrbt}[RRBT]{Rapidly Exploring Random Belief Trees}
    \acro{rgb}[RGB]{color}
    \acro{rgbd}[RGBD]{color-depth}
    \acro{rms}[RMS]{Root Mean Squared}
    \acro{rmse}[RMSE]{Root Mean Squared Error}
    \acro{ros}[ROS]{Robot Operating System}
    \acro{sde}[SDE]{Stochastic Differential Equation}
    \acro{SE3}[SE(3)]{Special Euclidean group in three dimensions}
    \acro{slam}[SLAM]{Simultaneous Localisation And Mapping}
    \acro{so3}[$\mathfrak{so}$(3)]{Lie algebra of special orthonormal group in three dimensions}
    \acro{SO3}[SO(3)]{Special Orthonormal rotation group in three dimensions}
    \acro{upm}[UPM]{Upsampled-Preintegrated-Measurement}
    \acro{uts}[UTS]{University of Technology, Sydney}
    \acro{vi}[VI]{Visual-Inertial}
    \acro{vio}[VIO]{Visual-Inertial Odometry}
    \acro{vo}[VO]{Visual Odometry}
\end{acronym}

\def\pc{{\mathcal{P}}}
\def\pt{{\mathbf{p}}}
\def\refframe{{\mathcal{F}}}
\def\time{{t}}
\def\map{{\mathcal{M}}}
\def\nn{{\mathcal{N}}}
\def\mean{{\mathbf{m}}}
\def\cov{{\text{cov}}}

\def\dscore{{s}}
\def\normal{{\mathbf{n}}}
\def\thr{{thr}}
\def\radius{{d_r}}
\def\voxelSize{{d_v}}
\def\nClouds{{N}}

\title{Dynamic Object Detection in Range data \\using Spatiotemporal Normals}
\author{Raphael Falque$^\dagger$, Cedric Le Gentil$^\dagger$, and Fouad Sukkar \\ Robotics Institute at the University of Technology Sydney \\ raphael.guenot-falque@uts.edu.au, cedric.legentil@uts.edu.au, fouad.sukkar@uts.edu.au  \\ $^\dagger$ Both authors contributed equally to this paper
}

\begin{document}

\maketitle




%

\begin{abstract}
On the journey to enable robots to interact with the real world where humans, animals, and unpredictable elements are acting as independent agents; it is crucial for robots to have the capability to detect dynamic objects. In this paper, we argue that the detection of dynamic objects can be solved by computing the spatiotemporal normals of a point cloud. In our experiments, we demonstrate that this simple method can be used robustly for LiDAR and depth cameras with performances similar to the state of the art while offering a significantly simpler method.
\end{abstract}

\section{Introduction}

\begin{figure}
    \centering
    \includegraphics[width=\linewidth]{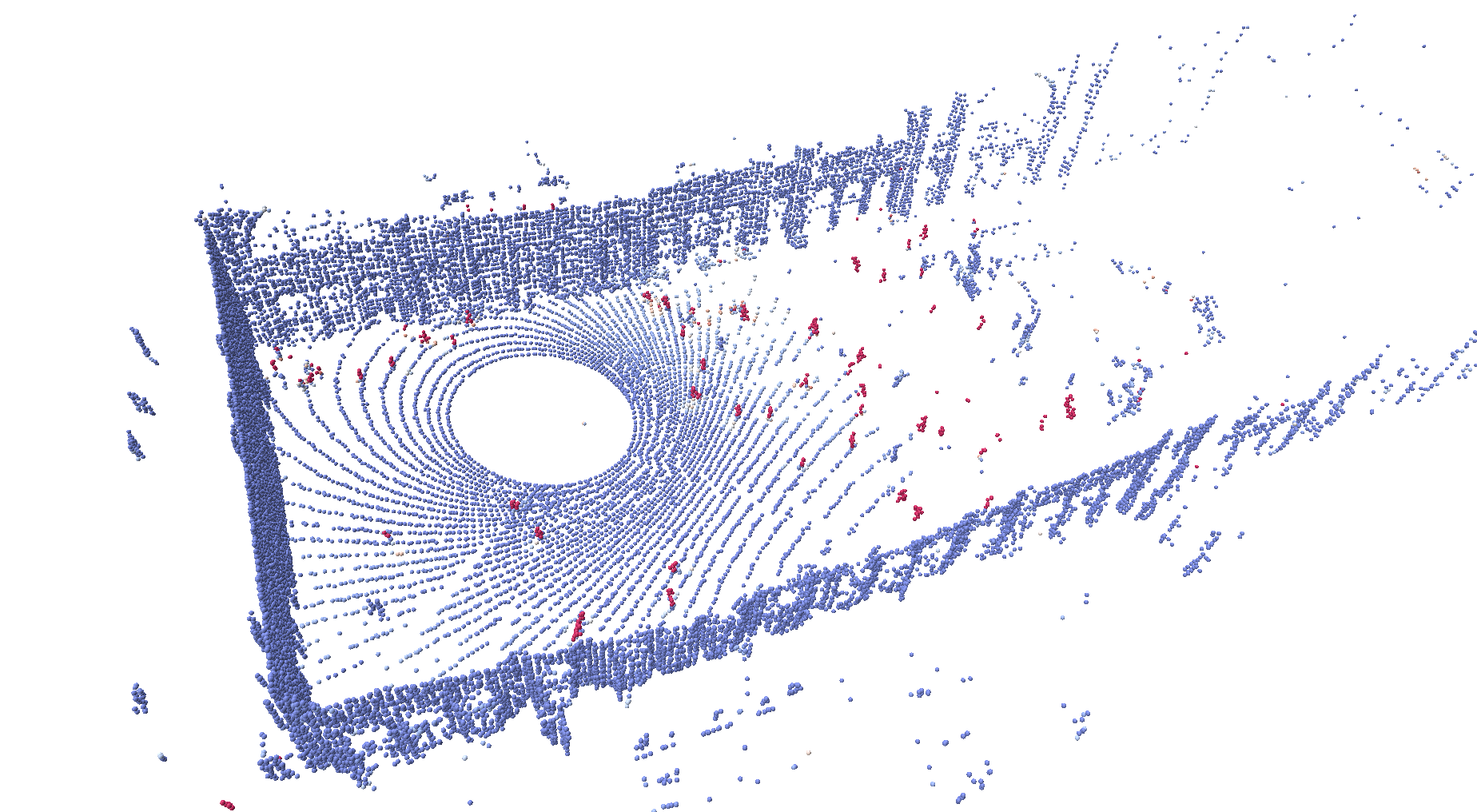}
    \caption{Illustration of the proposed method for detecting dynamic pedestrians in a large outdoor environment. The points that are detected as dynamic are shown in red and the static points are in blue.}
    \label{fig:teaser}
\end{figure}

Automation is a cornerstone of modern societies.
For example, one can think of the automatic production lines of most car manufacturers.
However, to this day, it is still rare to encounter autonomous robots outside of controlled environments such as large-scale factories, warehouses, etc.
A limiting factor is the ability of robots to adapt to uncontrolled situations or environments.
Such ability is crucial to ensure safe operation especially when humans are present in the environment.
Accordingly, as perception is the root of any autonomous system, the ability to detect dynamic objects in the surroundings of a robot is an essential step toward the democratisation of robots in society \cite{Cadena2016}.
Additionally, in the context of localisation and mapping, it has been shown that the detection of dynamic elements leads to nearly 40\% less odometry error \cite{pfreundschuh2021dynamic}.
As illustrated in Figure~\ref{fig:teaser}, this paper presents a method for dynamic object detection in data collected with range sensors such as lidars or \ac{rgbd} cameras.

With the rise of the deep learning dominance in the field of computer vision~\cite{krizhevsky2012imagenet}, the research community has naturally focused on the semantic segmentation of data as a proxy for classifying dynamic objects.
The most straightforward approach is to feed the sensor output into a 2D-\ac{cnn} and learn the dynamic components from large labelled datasets.
\cite{chen2019suma++} proposed to use such method with the \ac{cnn} Rangenet++~\cite{milioto2019rangenet++} and the KITTI Vision Benchmark dataset~\cite{geiger2012we}.
This work was later extended by stacking several consecutive scans in the network input and by benchmarking different \ac{cnn} architectures~\cite{chen2021moving}.
In \cite{Dai2018}, lidar data is upsampled into image-like data to fine-tune a \ac{cnn} object detector trained with standard images \cite{Redmon2017}.
As an alternative, in cases where the depth information is acquired with \ac{rgbd} cameras, the \ac{rgb} information can be used for the semantic segmentation and then transferred onto the map \ac{rcnn} architecture~\cite{henein2020dynamic}.


\begin{figure*}
    \centering
    \input{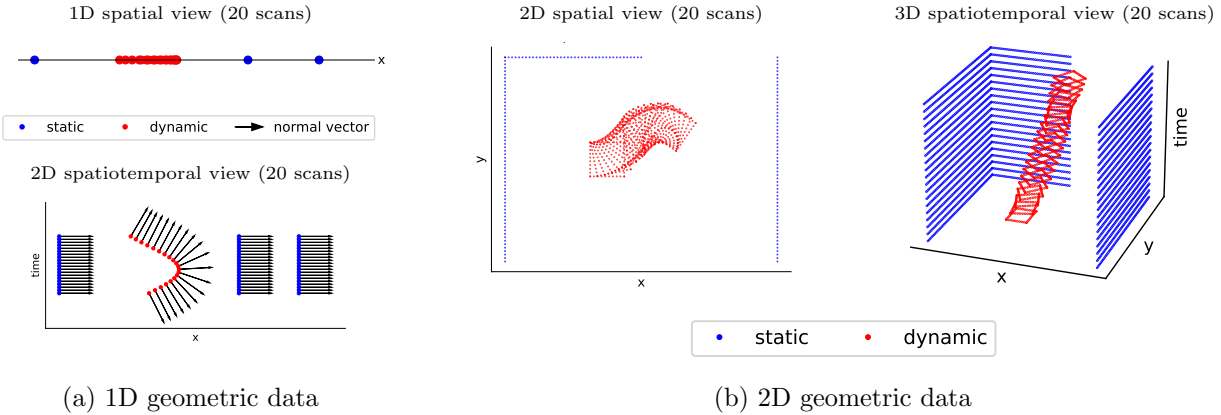}
    \caption{Spatiotemporal view of 1D (a) and 2D (b) geometric data in the presence of a dynamic object (20 range scans through time). It illustrates the link between the spatiotemporal normal vectors and the velocity of an object: for static objects the temporal component of the normals is null (the normals have been omitted in (b) for the sake of readability).}
    \label{fig:intuition}
\end{figure*}

While these methods allow the detection of dynamic objects, they are incapable of distinguishing moving objects from static objects such as parked cars. For this reason, a part of the literature is studying the detection of dynamic objects without consideration of the semantic information.

More recently, several approaches are trying to promote the idea of detecting dynamic objects in the map space in contrast to the sensor space. 
These approaches generally perform better as the map aggregates long-term information and the classification is reduced to finding areas that were once \emph{free} and are now \emph{occupied} (i.e., all the newly occupied areas are parts of dynamic objects).
A recent example of this strategy has been proposed by Schmid et al. where the Voxblox framework~\cite{oleynikova2017voxblox} has been extended for dynamic object detection~\cite{schmid2023dynablox}. 
Voxblox builds an efficient voxelization of the map with a hash map and Dynablox adds the mapping of the free-space confidence.
Similarly, Mersch et al. proposed to use a sparse 4D (i.e., [X, Y, Z, time]) \ac{cnn} to predict dynamic object and fuses the predictions of unoccupied space into the map with a Bayesian filter~\cite{mersch2023ral}.
Given the great performance of these occupancy-based approaches, they can be used as an offline tool for labelling large datasets which can later be used for the training of real-time deep learning architectures~\cite{pfreundschuh2021dynamic}.

In this paper, we propose a simple method based on the concept of spatiotemporal normals vectors for dynamic object detection.
The proposed method requires no training, has no requirement for any specific framework, and has a very limited number of parameters.
More precisely, the contributions of this work are as follows:
\begin{itemize}
    \item A simple yet effective method to detect dynamic objects in range data based on the analysis of the normal vectors in the spatiotemporal space.
    \item The open-source real-time implementation of our approach is available at \href{https://github.com/UTS-RI/dynamic_object_detection}{https://github.com/UTS-RI/dynamic\_object\_detection}.
\end{itemize}
Please note that the proposed method, similarly to \cite{schmid2023dynablox}, requires the input 3D scans to be already registered in a common referential frame.
The registration is outside the scope of this paper but can be performed simply using known kinematics if the sensor is mounted on a robot arm, or estimated with methods like \cite{LeGentil2021} or \cite{campos2023orbslam3} in the context of localisation and mapping.

\section{Method}

\subsection{Motivation and overview}

In this paper, we consider a stream of range data through time (e.g., from a lidar or a depth camera).
The goal is to classify the individual points of the incoming data as belonging to a dynamic object or not.
A dynamic object is defined as an object with a non-null velocity in an earth-fixed reference frame $\refframe_W$.
Accordingly, the proposed classification is agnostic to the nature of the objects present in the environment.

This work introduces a \emph{dynamic score} estimated for each of the incoming points based on the computation of spatiotemporal normal vectors.
In Figure~\ref{fig:intuition}, examples are provided to give an intuition on how the spatiotemporal normals relate to the points' velocities.
In the 1D scenario, the normal is directly linked to the velocity.
In higher dimensions, the temporal component of the spatiotemporal normal corresponds to the projection of the actual velocity on the spatial normal to the surface.

\begin{figure*}
    \centering
    \def\hdist{4em}
    \begin{tikzpicture}[auto]
        \tikzstyle{input} = [draw, fill=white, rectangle, minimum height = 2.4em, text width = 5.5em,  minimum width = 5.5em, align = center, node distance = 10em, draw=red, execute at begin node=\setlength{\baselineskip}{8pt}]
        \tikzstyle{block} = [draw, fill=white, rectangle, minimum height = 6.4em, text width = 8.5em,  minimum width = 8.5em, align = center, node distance = 11em, execute at begin node=\setlength{\baselineskip}{8pt}] 
        \tikzstyle{output} = [draw=blue, fill=white, rectangle, minimum height = 3em, text width = 8.5em,  minimum width = 8.5em, align = center, node distance = 11em, execute at begin node=\setlength{\baselineskip}{8pt}] 

        \node [input] (rawdata) {\scriptsize \textbf{Range data} (lidar or depth camera)};
        \node [block, right of=rawdata, xshift = -1.5em] (registration) {\scriptsize \textbf{Point cloud registration}\\Odometry, motion distortion correction, robot arm kinematics, etc};
        \node [block, right of=registration ] (nn) {\scriptsize \textbf{Local map $\map$ and nearest neighbours}\\Point cloud aggregation (temporal neighbourhood) and per-point spatial neighbourhood search};
        \node [block, right of=nn] (pca) {\scriptsize \textbf{Spatiotemporal normal computation}\\Local covariance computation, Eigendecomposition};
        \node [output, right of=pca] (score) {\scriptsize \textbf{Dynamic score and classification}\\Temporal component of spatiotemporal normal as dynamic score $\dscore_i^j$, thresholding};
        \draw[->] (rawdata) -- (registration);
        \draw[->] (registration) -- node{\scriptsize$\pc_i$} (nn);
        \draw[->] (nn) -- node{\scriptsize $\nn_i^j$ }(pca);
        \draw[->] (pca) -- node{\scriptsize $\normal_i^j$ }(score);
        \coordinate[below= \hdist of score.east] (bracketa);
        \coordinate[below= \hdist of nn.west] (bracketb);
        \draw [very thick,decorate,decoration={calligraphic brace,amplitude=10pt}] (bracketa)  --(bracketb);
        \coordinate (mid) at ($(bracketa)!0.5!(bracketb)$);
        \node[below=1.5em of mid]{Proposed method};
        \begin{scope}[auto]
            \def\legendx{-0.5}
            \def\legendy{-2.0}
            \node[rectangle, draw=red] at (\legendx,\legendy) {};
            \node[anchor=west] at (\legendx+0.2,\legendy) {Input};
            \node[rectangle, draw=blue] at (\legendx+2.5,\legendy) {};
            \node[anchor=west] at (\legendx+2.7,\legendy) {Output};
        \end{scope}

    \end{tikzpicture}
    \caption{Overview of the proposed method for dynamic object detection in range data. The method relies on registered point clouds aggregated in local maps. The final \emph{dynamic score} relies on the computation of the spatiotemporal normal vector for each of the points.}
    \label{fig:overview}
\end{figure*}
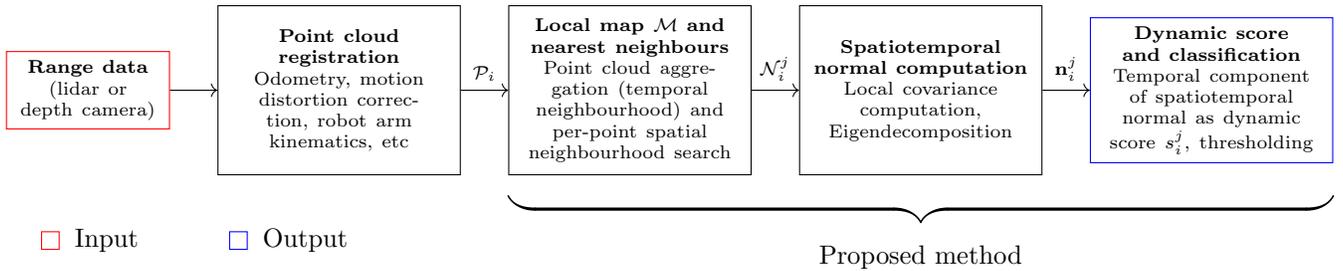

An overview of the proposed pipeline is shown in Figure~\ref{fig:overview}.
The input of our method consists of point clouds registered in a fixed reference frame.
In the context of mapping, the registration can be performed based on odometry information (lidar, visual, etc.) while the kinematics of a robot arm can be leveraged for cobotics-manipulation scenarios.
Similarly to standard normal vector estimation, the spatiotemporal normals are computed through a \ac{pca} of the point's neighbourhood.
More specifically, the Eigendecomposition of the local covariance of the data is performed and the Eigenvector corresponding to the smallest Eigenvalue is used as the normal estimate. 
As discussed above, such a normal vector in the spatiotemporal space provides information regarding the point's velocity.
The dynamic score is then defined as the time component of the spatiotemporal normal.
The following subsection presents a formal derivation of the proposed dynamic score.

\subsection{Dynamic score via spatiotemporal}

 Let us consider an undistorted point cloud $\pc_i$ ($i\in1,\cdots,N)$ registered in a unique fixed frame $\refframe_W$.
 We denote $\pt_i^j$ the j$^{th}$ point in $\pc_i$ and $\time_i^j$ the associated timestamp.
 The symbol $\map$ refers to the aggregation of all the point clouds $\pc_i$ so that $k \leq i \leq l$ (with $k$ and $l$ derived from the method's parameters).
To provide a dynamic score to a point $\pt_i^j$, we first perform a neighbourhood search in $\map$.
We define $\nn_i^j$ as the set of neighbour points.
The local covariance $\cov_i^j$ is computed as
\begin{equation}
    \cov_i^j = \frac{1}{\Vert\nn_i^j\Vert}\sum_{\pt_u^v \in \nn_i^j} \left(\begin{bmatrix}\pt_u^v\\ \time_u^v\end{bmatrix} - \mean_i^j\right)\left(\begin{bmatrix}\pt_u^v\\ \time_u^v\end{bmatrix} - \mean_i^j\right)^\top
\end{equation}
with
\begin{equation}
    \mean_i^j = \frac{1}{\Vert\nn_i^j\Vert}\sum_{\pt_u^v \in \nn_i^j} \begin{bmatrix}\pt_u^v\\ \time_u^v\end{bmatrix}.
\end{equation}

The dynamic score $\dscore_i^j$ is defined as the absolute value of the temporal component of the Eigenvector associated with the smallest Eigenvalue of $\cov_i^j$.
A low dynamic score corresponds to low velocity in the 3D space.
Thus, the point classification as static or dynamic is performed by thresholding $\dscore_i^j$.

\subsection{Implementation}
The proposed method has been implemented to run in real-time in C++ and written as a \ac{ros} node that \emph{subscribes} to a point cloud $\pc_{in}$ and \emph{publishes} two point clouds, one dynamic $\pc_{out}^{dyn}$ and one static $\pc_{out}^{sta}$.

The implementation first performs a voxel grid downsampling, with a voxel size $\voxelSize$, using a hash map~\cite{niessner2013real} and aggregates the point clouds in the map $\map$.
The map is defined as a sliding window over the last $2\nClouds+1$ point clouds, where similarly to a \ac{fifo} pile, each time a new point cloud is aggregated into $\map$, the oldest point cloud is then removed.
A kd-tree~\cite{blanco2014nanoflann} is built with $\map$ and for each of the points in the $(\nClouds+1)^{th}$ last point cloud, a query of the neighbours within the radius $\radius$ is then performed to compute the spatiotemporal normal using \ac{pca}.
The dynamic points are then detected by using a threshold $\thr$ over the temporal component of the computed normal.

Eventually, our method relies on only four parameters: the voxel size $\voxelSize$, the number of clouds $\nClouds$, the radius for the neighbour search $\radius$, and the dynamic score threshold $\thr$. All these parameters have a physical meaning and are easy to tune depending on the application requirements. 
The parameters on the voxel size, $\voxelSize$, and the number of aggregated point clouds, $\nClouds$, have a direct relationship with the computational time and can be used to balance sensors with high resolution and high framerates. 
The radius used for finding the points' neighbours, $\radius$, has a marginal impact on the computational time and should be chosen with respect to the scale of the scanned environment.
Finally, the threshold, $\thr$, should be chosen depending on how important it is to remove dynamic points.

\section{Experiments}

The experiment section is separated into two parts, first, we test the proposed method on an established dataset which has been used consistently in the recent state of the art~\cite{pfreundschuh2021dynamic,schmid2023dynablox}, secondly, we show a practical study case on how the dynamic coefficient can be used in human-robot collaboration scenario.
Different parameters are used for each scenario as the experiments are performed with sensors that have radically different ranges (LiDAR and \ac{rgbd} camera).

\subsection{The DOALS dataset}

\begin{figure*}
    \centering
    \begin{subfigure}[t]{0.3\linewidth}
        \centering
        \includegraphics[width=\linewidth]{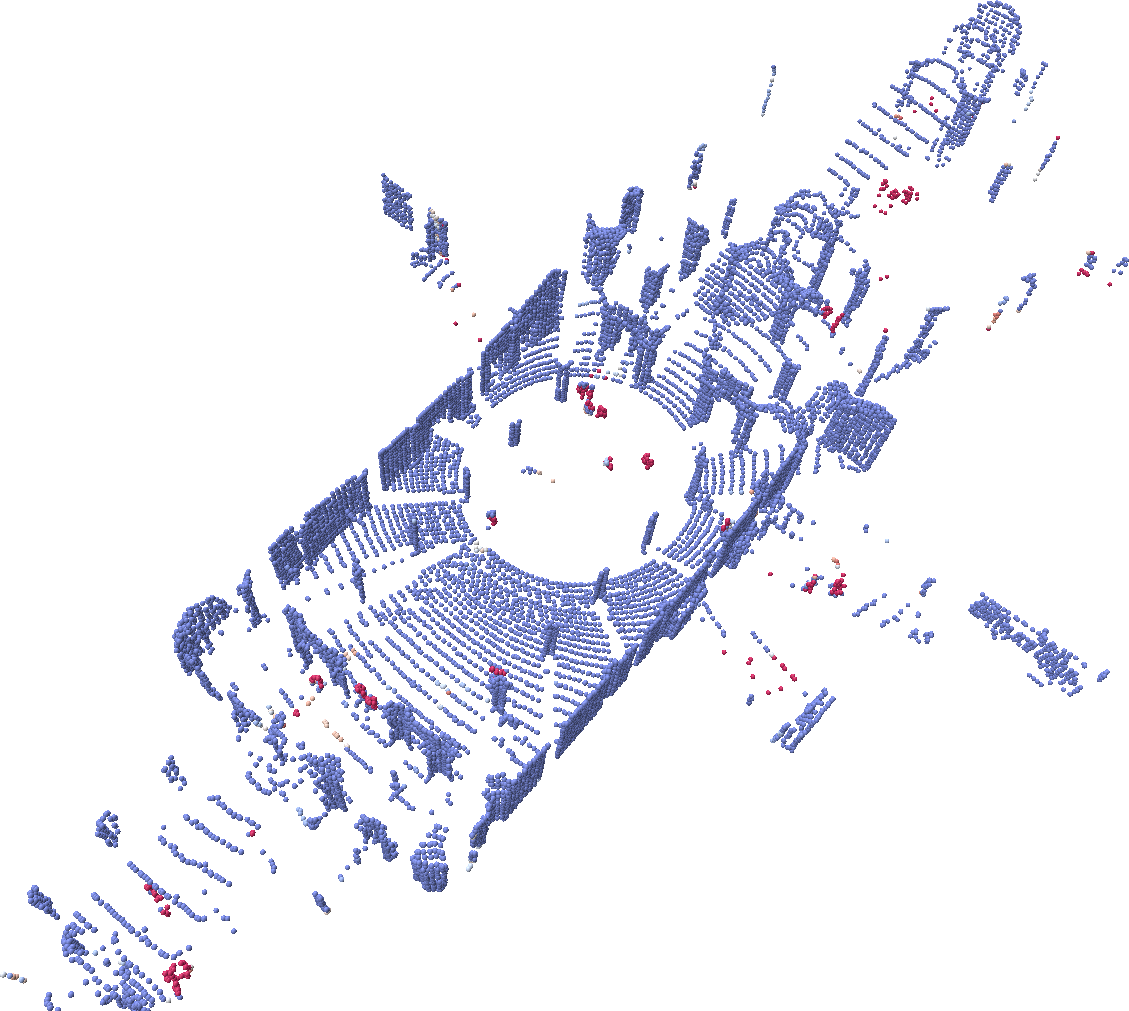}
        \caption{Proposed method}
    \end{subfigure}
    \hfill
    \begin{subfigure}[t]{0.3\linewidth}
        \centering
        \includegraphics[width=\linewidth]{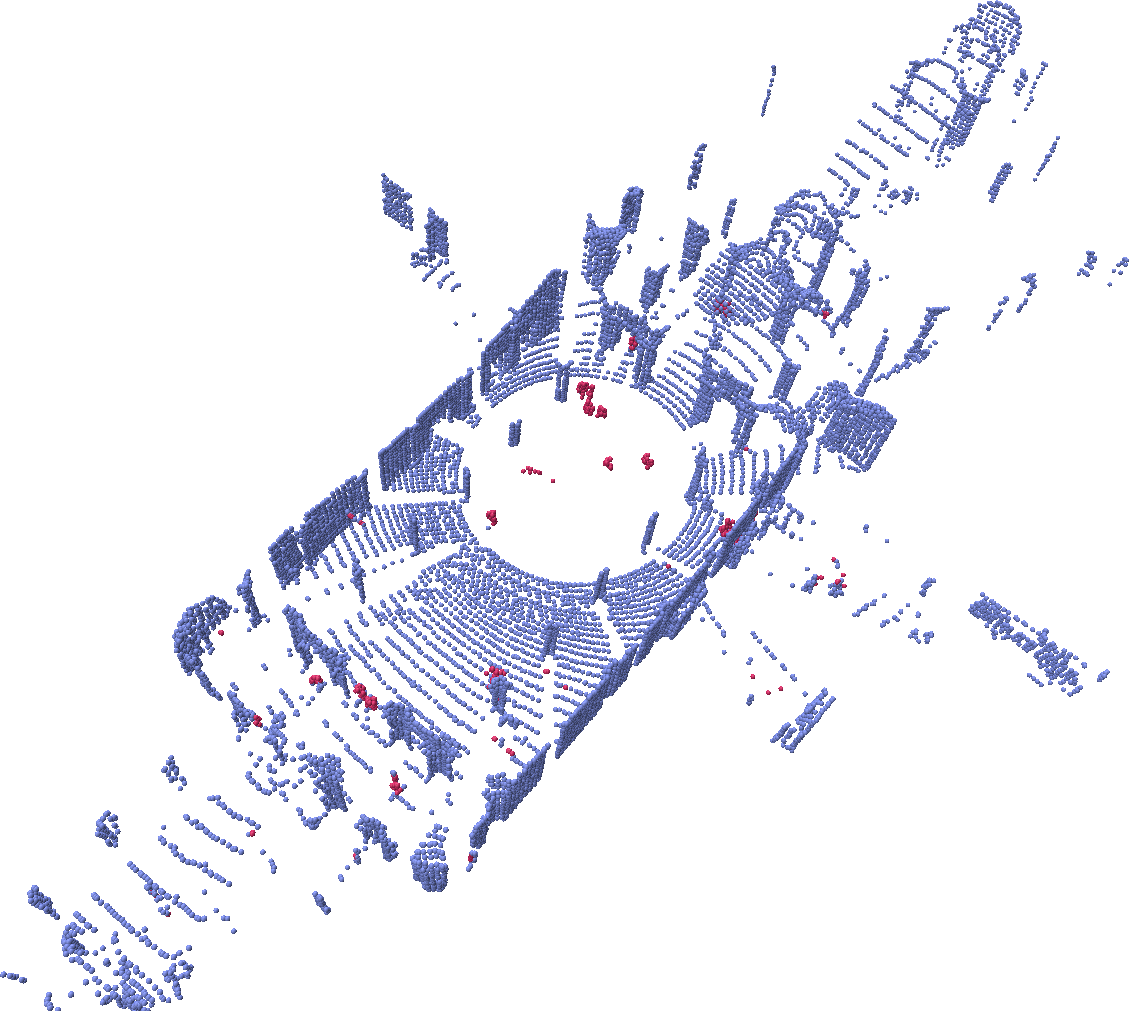}
        \caption{Dynablox~\protect\cite{schmid2023dynablox}}
    \end{subfigure}
    \hfill
    \begin{subfigure}[t]{0.3\linewidth}
        \centering
        \includegraphics[width=\linewidth]{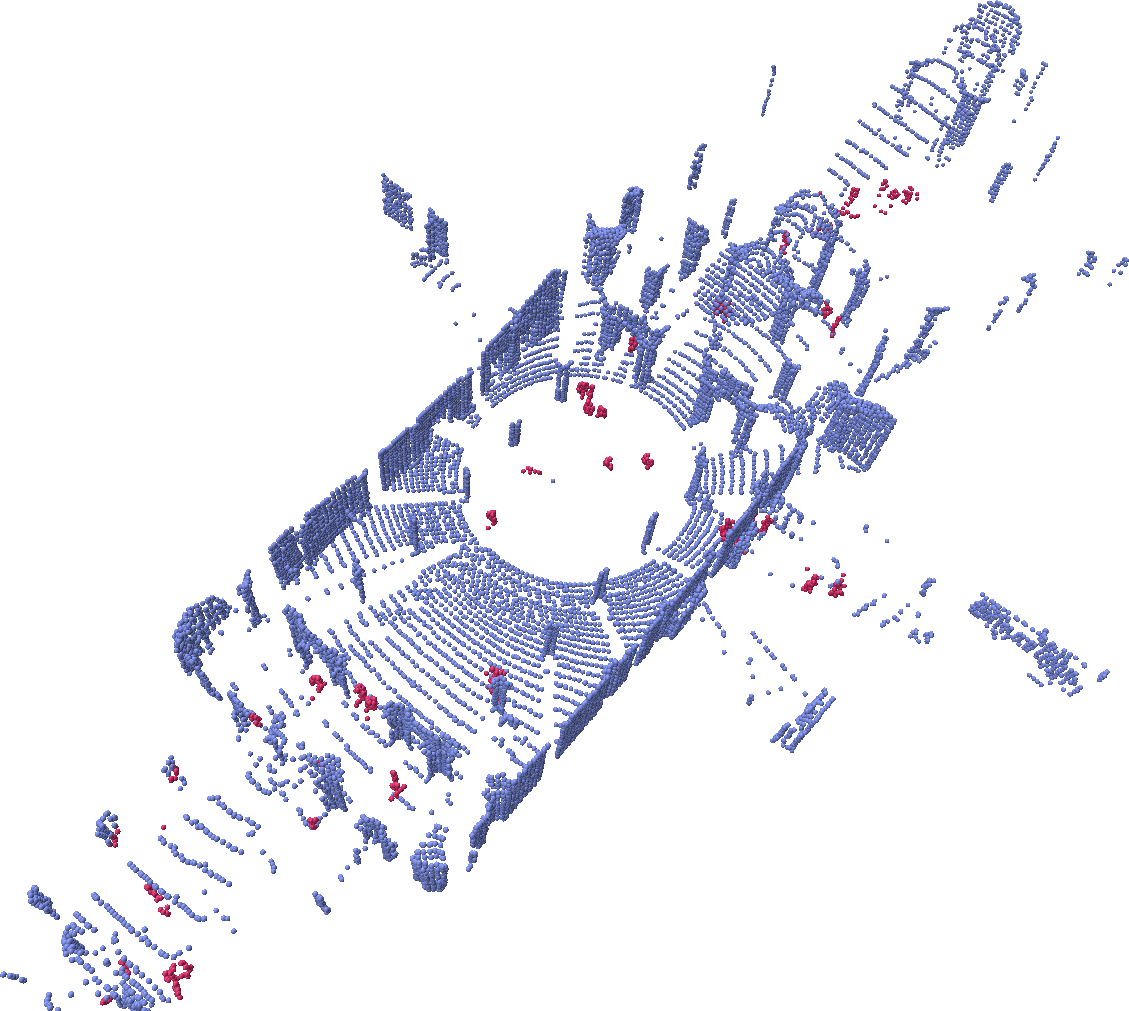}
        \caption{Ground truth}
    \end{subfigure}%
    \vspace{-0.2cm}
	\caption{Comparison between the proposed method (a), Dynablox (b) and the ground truth labels (c) on the \emph{Hauptgebaeude} scene from the Urban Dynamic Objects LiDAR (DOALS) Dataset~\protect\cite{pfreundschuh2021dynamic}.}
	\label{fig:DOALS}
\end{figure*}

We applied our method on the \ac{doals} dataset~\cite{pfreundschuh2021dynamic}. 
The dataset contains two scenarios: the first one is made of a simulated small town with moving elements (e.g., cars, planes, pedestrians, animals, etc.) that provide proper ground truth for dynamic elements; the second scenario is from real-world scenes recorded around the Zurich metropolitan area. 
An Ouster OS1 64 LiDAR, producing clouds with $131.072$ points at 10 Hz, was used for the data collection and manual annotation was provided for some of the timestamps. 
In this experiment, we use the following parameters: $\nClouds=10$, $\radius=0.3$, and $\thr=0.25$ and the voxel size $\voxelSize$ is set proportionally to the scale of the point cloud (i.e., the diagonal of its bounding box) as follows:
\begin{equation}
    \voxelSize = \frac{\text{scale}(\pc_i)}{600}.
\end{equation}
A qualitative evaluation of our method is shown in Figure~\ref{fig:DOALS} where we compare the dynamic coefficient to the prediction of Dynablox and the ground truth annotations.

Furthermore, to provide a quantitative evaluation, we compare the \ac{iou} for LC Free space~\cite{modayil2008initial}, Dynablox, and the proposed method in Table~\ref{tab:quantitative_results} on all real scenes from the \ac{doals} dataset\footnote{We did not manage to run Dynablox on the virtual dataset.}. 
The evaluation is performed on the original point cloud resolution for Dynablox and LC Free Space, while we upsample our dynamic prediction to the original point cloud resolution. The upsampling from the downsampled cloud to the full-resolution cloud is performed by a local search in the vicinity of the dynamic points.
More specifically, for all dynamic points in the downsampled cloud, we search for their neighbours within a radius of 0.5m in the full-resolution cloud and define them as dynamic.

\begin{table}
    \centering
    \caption{Quantitative evaluation of the Intersection over Union (IoU) [\%] between LC Free space~\protect\cite{modayil2008initial}, Dynablox~\protect\cite{schmid2023dynablox}, and the proposed method. A limit is applied to the range at 20m. SV, HG, and ND stand for the Shopville, Niederdorf, and Hauptgebaeude scenes respectively. We did not manage to reproduce exactly the same performances as the original paper for Dynablox which were slightly higher.}
    \begin{tabular}{ l c c c c c }
     \toprule
                &  Station  &  SV    &  HG   & ND    & all   \\ 
     \midrule
     LC Free space &   0.49     &  0.32         &  0.25    &  0.18            & 0.31     \\ 
     Dynablox   &   0.81     &  0.84         &  0.82    &  0.81            & 0.82     \\ 
     ours       &   0.80     &  0.81         &  0.85    &  0.76            & 0.81     \\ 
     \bottomrule
    \end{tabular}
    \label{tab:quantitative_results}
\end{table}

The results from Table~\ref{tab:quantitative_results} show that the proposed method has performances relatively similar to the state of the art while alleviating the requirement for complex frameworks. 

\subsection{Study case: human robot interaction}

Unlike traditional robotic arms, collaborative robots (cobots) are designed to be operating with humans in their surroundings.
There are typically four types of cobot \ac{hri}: physically separated, coexistence, cooperation, and collaboration~\cite{guertler2023robot}. 
For the latter three cases, the cobot must react and adapt to the human to safely carry out tasks. 
In the case of coexistence and cooperation, a common control strategy is to slow down the cobot based on its proximity to any humans~\cite{marvel2013performance,villani2018survey}.
For collaboration, our method could be used in conjunction with dynamic motion planning algorithms~\cite{likhachev2005anytime,ferguson2006using,alwala2021joint} for detecting changes in the environment and enabling the cobot to react in real-time to these changes.

\begin{figure}
    \vspace{-0.5cm}
    \centering
    \def\vertdist{0.45cm}
    \def\subdist{-0.05cm}
    \def\horidist{0.0cm}
    \def\photoscale{0.62}
    \def\scanscale{0.35}
    \def\subsize{\small}
    \begin{tikzpicture}
        \node (photo){\includegraphics[width=\photoscale\columnwidth]{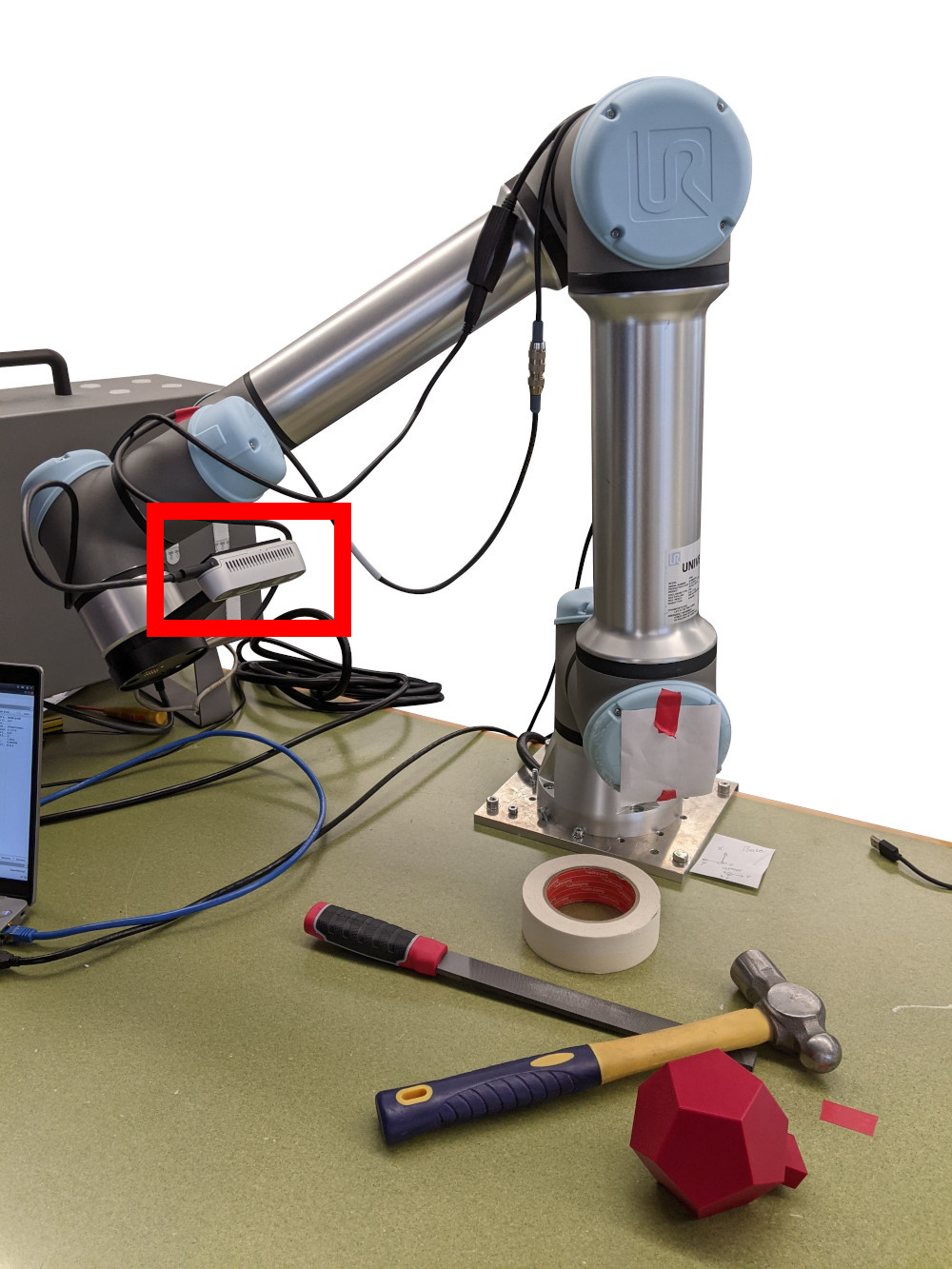}};
        \node[right=\horidist of photo.south east, anchor=south west] (score) {\includegraphics[clip, trim=0cm 0.5cm 0cm 0cm, width=\scanscale\columnwidth]{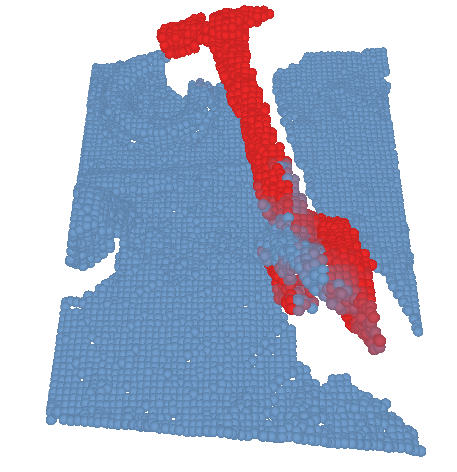}};
        \node[above=\vertdist of score] (raw)                                 {\includegraphics[clip, trim=0cm 0.5cm 0cm 0cm, width=\scanscale\columnwidth]{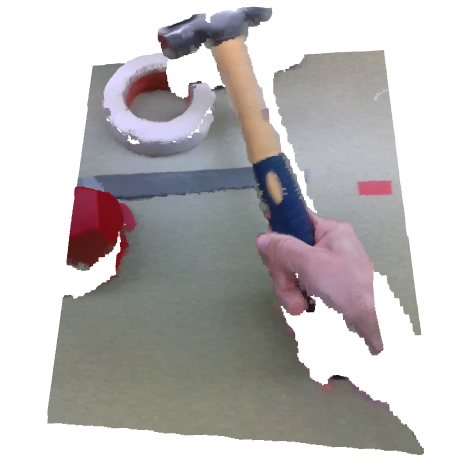}};
        \node[below=\subdist of score] {\subsize(c)};
        \node[below=\subdist of raw] {\subsize(b)};
        \node[below=\subdist of photo] {\subsize(a)};
    \end{tikzpicture}
    \vspace{-0.8cm}
    \caption{\ac{hri} experiment with a shared workspace between a human operator and a robot arm. The real-time detection of the dynamic motion can help the robot to avoid harming the human operator. The shared workspace is shown in (a), with the scanned objects, the robot arm, and the mounted \ac{rgbd} Realsense camera. The capture \ac{rgbd} data are displayed in (b), and the dynamic classification of the points is shown in (c) with the dynamic points in red and the static points in blue.}
    \label{fig:study_case}
\end{figure}
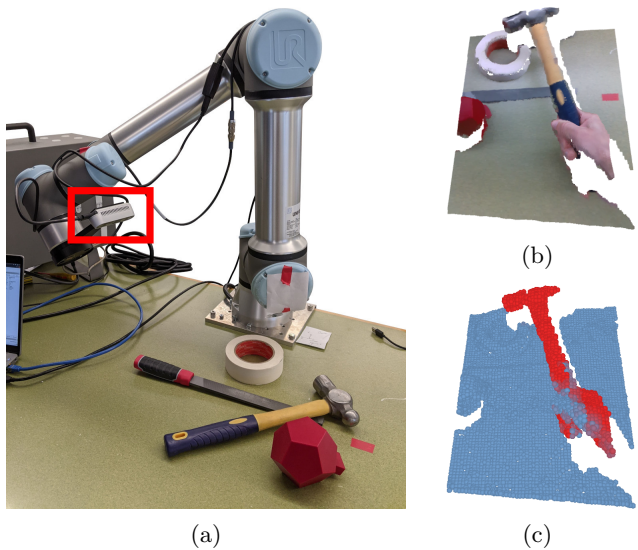

In this study case, we consider a shared workspace between a robot arm and a human such as the one shown in Figure~\ref{fig:study_case}(a).
We demonstrate how an embedded perception system can allow a robot arm to be \emph{aware} of its surroundings by detecting moving elements in the workspace.
Our experimental setup consists of a UR5 robot arm equipped with an \ac{rgbd} Intel Realsense d435 camera on its end effector.
The arm moves according to randomly selected 6-DoF waypoints above a table containing multiple objects.
While the robot is moving, a human operator displaces some of the objects in the workspace.

The \ac{rgbd} camera point clouds are projected into the robot's base referential frame according to the arm's kinematics.
An example of raw data and estimated dynamic score are shown in Figure~\ref{fig:study_case}(b) and (c).
The parameters for this experiment are set as $\nClouds=4$, $\radius=0.3$, $\thr=0.01$ and $\voxelSize=\frac{\text{scale}(\pc_i)}{100}$.
Additionally, the real-time detection of the moving elements is showcased in the attached video where both the registered point clouds and their associated dynamic scores are displayed.

\section{Limitations}
One of the major limitations of the proposed approach is the difficulty of detecting dynamic points close to the ground as part of the ground is picked in the neighbour search and changes the direction of the spatiotemporal normal. 
Such a problem could be avoided by removing the ground which is a common approach in the literature~\cite{petrovskaya2009model,postica2016robust,arora2021mapping,arora2023}.

In the current implementation, one of the main limitations for safety applications consists of the delay between the dynamic predictions and the data stream. More specifically, given a sensor that generates depth measurements with a frequency $f$, the delay is $\dfrac{\nClouds}{f}$ seconds. This limitation can be mitigated by increasing the framerate from the sensor or by limiting the number of clouds used for the dynamic score prediction.

\section{Conclusion}
In this paper, we propose a simple yet effective method for detecting dynamic elements.
Our method is on par with the state of the art in terms of performance while still keeping an extremely simple formulation which is a guarantee for robustness.
Therefore thanks to its simplicity we do believe that the proposed approach can be used as a stepping stone for more complex frameworks. 
A typical application could be to use the dynamic score in the feature space of learning algorithms for end-to-end frameworks.

In future work, we are planning to explore the coupling of the dynamic perception system with the control part of a robot arm, to provide a safer and \emph{aware} interaction for \ac{hri}. Furthermore, we want to integrate the dynamic object detection into a full \ac{slam} framework for the generation of clean maps.

\section{Acknowledgment}

Cedric Le Gentil is supported by the Australian Research Council Discovery Project under Grant DP210101336.

\bibliographystyle{named}
\bibliography{bibliography}
\balance

\end{document}